\pdfoutput=1

\documentclass[11pt]{article}

\usepackage[]{naacl2021}

\usepackage{times}
\usepackage{latexsym}

\usepackage[T1]{fontenc}

\usepackage[utf8]{inputenc}

\usepackage{microtype}
\usepackage{multirow}
\usepackage{graphicx}
\usepackage{amsmath}
\usepackage{soul}
\setuldepth{Berlin}

\title{Detecting over/under-translation errors for determining adequacy in human translations}

\author{Prabhakar Gupta\thanks{~Both had equal contribution} \\ prabhgup@amazon.com \And
        Ridha Juneja\footnotemark[1] \\ ridhaj@amazon.com \And
        Anil Nelakanti \\ annelaka@amazon.com \And
        Tamojit Chatterjee \\ tamoji@amazon.com}
  
\begin{document}
\maketitle
\begin{abstract}

We present a novel approach to detecting over and under translations (OT/UT) as part of \textit{adequacy} error checks in translation evaluation.
We do not restrict ourselves to machine translation (MT) outputs and specifically target applications with human generated translation pipeline.
The goal of our system is to identify OT/UT errors from human translated video subtitles with high error recall. We achieve this without reference translations by learning a model on synthesized training data.
We compare various classification networks that we trained on embeddings from pre-trained language model with our best hybrid network of GRU + CNN achieving  89.3\% accuracy on high-quality human-annotated evaluation data in 8 languages.

\end{abstract}

\section{Introduction}
Quality evaluation (QE) in machine translation (MT) focuses on evaluating translation quality without reference text~\cite{DBLP:conf/coling/BlatzFFGGKSU04, DBLP:conf/eamt/SpeciaTCCD09}.
Translation quality captures both the \textit{fluency} of the translation and its \textit{adequacy} relative to the source. 
Given a source sentence $s$ and corresponding MT output $t$, QE system is a mapping~$f:(s,t)\rightarrow r$ quantifying translation quality.
Human-mediated Translation Edit Rate (HTER)~\cite{Snover2006ASO} is often used for $r$ to quantify the human post-edit burden to arrive at good translations.\footnote{Note that one needs reference translation to compute HTER while most QE systems attempt to estimate them in the absence of references.}
Recent advancements in QE modelling have attempted with limited corpora that surfaced numerous undesirable artefacts. 
For example, \cite{openkiwi} and \cite{DBLP:journals/corr/abs-2004-13828} model QE as a classification problem that grades the semantic match between original source and the machine translated target.
They provide a binary verdict of the translation as good/bad with very low miss-rate leaving the final judgements to translators. 
It is unclear if a monolith binary classifier can learn to detect multiple errors of very different nature with varying syntactic and semantic language patterns. 
Due to issues like sampling bias and lexical homogeneity in data, pre-trained models could game the tasks without actually learning to evaluate quality ~\cite{DBLP:conf/acl/SunGS20}. 
Further, considering usability, a good/bad label is less helpful to post-editors than fine-grained errors that mark a pair as, say, over-translation likely due to these tokens in target.

While human translations (HT) are synonymous to gold standard for purposes of MT system evaluations, numerous operational pipelines that employ humans for translating often return output with much lower than acceptable quality.
Relaxing the scope of QE to include HT can help assisting humans translators like various applications do for spell checkers and English grammar.
Digital entertainment industry, for example, makes extensive use of professional human translators for translating video subtitles to increase content viewership. Translated subtitles often require human quality checks that are as expensive as acquiring translations~\cite{DBLP:journals/jides/Poirier14}.
To reduce post-editing quality checks costs, we could flag errors as the translations are typed in with the QE serving as a guardrail.

However, is such broadening of scope for QE to identify error patterns in any generic translation possible? 
In this work, we work towards this direction addressing specifically over-/under-translations (OT/UT) that constitute one subset of adequacy errors.
We propose to learn classification models that directly evaluate fluency and adequacy without resorting to HTER estimates as an intermediate step.
In addition to evaluating quality, this approach offers editing cues to human post-editors by flagging specific errors.
%
%
%
Each translation pair $(s,t)$ is labelled as either OT or UT or otherwise as good class with no error (NE). OT (or UT) captures the asymmetry with source containing less (or more) information than the target~\footnote{For detailed description of errors refer to ~\cite{DBLP:journals/corr/abs-1909-05362} and Multidimensional Quality Metrics (MQM) (www.qt21.eu/mqm-definition/definition-2015-12-30.html)}. 
We use pre-trained language models (LM) to synthesize training data and train a binary classifier using synthetic data.
We experimented with various choices for classifier heads that are compared in the results section.
Evaluation uses human annotators to validate model predictions across different languages.

This problem becomes much more difficult for video subtitles because it is possible for a translation to be linguistically incomplete and be acceptable during post-edits. This is due to the fact subtitles are required to follow a set of technical constraints limiting the choice and number of words in translation. For example, an English subtitle: \textit{``There is a green tree in the park''} can be translated to \textit{``Green tree in park''} and still be considered acceptable because it is successfully conveying the context to the viewer. 

Considering this idiosyncratic behaviour of video subtitles, we present novel approaches to generate synthetic training data that capture errors frequently encountered in human translations. 
Our focus is on improving operational efficiency of pipelines involving human translators and we specifically focus on subtitling on which we evaluate our methods.
We show the effectiveness of our model in detecting OT and UT in human translations without any access to reference translations. 
The goal is to operate at high error recall to ensure poor subtitles are not pushed into the final data to be viewed by numerous customers.
Finally, we compare the performances from four of our networks with the hybrid GRU and CNN model performing best on our high-quality evaluation data acquired from multiple human translators.

\begin{table*}[h]
  \centering
  \caption{English-French subtle negatives for UT}
    \begin{tabular}{p{0.31\textwidth}|p{0.31\textwidth}|p{0.31\textwidth}}
    \textbf{Original Source Subtitle} & \textbf{New Source Subtitle} & \textbf{Target Subtitle} \\ \hline
    it is my duty to remind you of what you've got there. & it \ul{still} is my duty to remind you \ul{fully} of what you've \ul{already} got \ul{done} there \ul{recently}. & il est de mon devoir de te rappeler ce que tu as. \\ \hline
    put that book away for a while to make money, & put that book away for a \ul{good} while to make money, & mettre ton livre de côté pour gagner un peu d'argent, \\ \hline
    
    I didn't know if you'd have your luggage with you. I have no luggage. & I didn't know if \ul{even} you'd \ul{always} have your luggage with you. I have no luggage. & J'ignorais si vous aviez des bagages. Je n'ai pas de bagage. \\ 

    \end{tabular}%
    \label{tab:under-translation-soft}%
\end{table*}

\begin{table*}[h]
  \centering
  \caption{English-French subtle negatives for OT}
    \begin{tabular}{p{0.31\textwidth}|p{0.31\textwidth}|p{0.31\textwidth}}
    \textbf{Original Source Subtitle} & \textbf{New Source Subtitle} & \textbf{Target Subtitle} \\ \hline
    Please \ul{take} care of \ul{Espada} until this war is over. & Please care of until this war is over. & Prenez soin d'Espada en attendant que la guerre soit finie. \\ \hline
    We were \ul{gonna} \ul{stop} at the Elephant \ul{Cafe}. & We were at the Elephant. & Je veux dire, on allait s'arrêter à l'Elephant Cafe. \\ 
    
    \end{tabular}%
    \label{tab:over-translation-soft}%
\end{table*}

\section{Related Work} 
There is considerable work on QE for MT~\footnote{see~\cite{DBLP:conf/wmt/FonsecaYMFF19} for a review of various methods}.
Fitting discriminative models to hand-crafted linguistic features trained to estimate HTER was common until pre-trained neural models with regression heads came along with better metrics.
MT output with reference translations are necessary to train these models which will then estimate the likely HTER for any given $(s,t)$ translation pair.
Correspondingly, error patterns are restricted to those from MT systems used for generating the data.

In expanding error patterns beyond MT, we will need alternate approaches for generating training data that capture various plausible adequacy errors.
Also, gathering reference texts is a blocker in scaling the approach across both, error patterns and languages.
Synthesizing errors to generate negatives for the model to learn from is a suitable alternative where feasible.
\cite{popovic2011towards}, for example, use linguistic features to synthesize errors for MT output analysis.
This approach is common in learning grammatical error checkers that uses syntactic structure to generate errors for model training  (see~\cite{DBLP:conf/bea/OmelianchukACS20, DBLP:conf/acl/KanekoMKSI20, DBLP:journals/tacl/LichtargeAK20}).

We follow a similar approach to synthesize errors to train our models.
The closest to our work is that of \citet{DBLP:conf/acl/TuLLLL16} that describes a coverage mechanism for detecting OT and UT errors in neural MT outputs. It has two drawbacks; they require reference text to generate the coverage matrix and their definition of OT/UT errors is also limited to source-side errors ignoring target-side additions and omissions.
\cite{DBLP:conf/nlpcc/YangZQZLS18} attempts to improve the BLeU scoring metric \cite{DBLP:conf/acl/PapineniRWZ02} by handling OT/UT errors but again they rely on reference translations for their evaluations.

We give details of the methods we employ to generate subtle and gross errors for OT and UT across languages using a pre-trained LM followed by a discussion of our evaluation data, model architectures, experiments and their results.

\begin{table*}[h]
  \centering
  \caption{English-French gross negatives for OT and UT}
    \begin{tabular}{p{0.31\textwidth}|p{0.31\textwidth}|p{0.31\textwidth}}
    \textbf{Original Source Subtitle} & \textbf{New Source Subtitle} & \textbf{Target Subtitle} \\ \hline
    Ivan is set to pull out of this place in a week.\ul{ We're moving to Antigua.} & Ivan is set to pull out of this place in a week. & Ivan a organisé son départ dans une semaine. - On s'installe à Antigua. - Et? \\ \hline
    - Fair enough. - So? & \ul{People had envisioned a monster.} - Fair enough. - So? & - C'est juste. - Alors ? \\
    \end{tabular}%
    \label{tab:hard-candidates}%
\end{table*}

\section{Data generation}

We began with a set of fairly clean subtitle files appearing on movies and TV shows from a subscription video-on-demand provider. We restricted the source language to English and target to be in one of 28 languages listed below.
All sentences in our set were between $5$ and $60$ tokens long and the sentence-pair cosine similarity score was $>0.8$ when computed using LASER embeddings~\cite{DBLP:journals/tacl/ArtetxeS19}.
This ensured that the seed set of 20MM subtitle translations we began with were a fairy-clean sample with good translation pairs.

Synthesizing OT/UT errors requires adding and omitting words to create information asymmetry between $s$ and~$t$.
We modify only English language source sentence $s$ for both OT/UT errors by omitting/adding words and rely on pre-trained multi-lingual models to handle cross-lingual patterns~\cite{DBLP:conf/emnlp/WuD19}.
This is so that we could leverage various openly available English language tools for sentence edits.

\textbf{UT subtle negatives.} For UT subtle negatives, we inserted tokens in source sentence $s$ chosen by probing a pre-trained English BERT LM~\cite{DBLP:conf/naacl/DevlinCLT19} with mask at a randomly chosen index. We filter out token suggestions with subwords, punctuation, stop words, special symbols, previous/next word repetitions and numerals.
We insert up to five tokens incrementally rather than filling multiple tokens at once as that could yield meaningless candidates. 
To filter out edits with minimal changes we compute a similarity score of source $s$ and candidate $s'$ using cosine of averaged Glove word embeddings~\cite{DBLP:conf/emnlp/PenningtonSM14}.
Top $40^{th}$ percentile constituting the most similar candidates sorted by this similarity score are filtered out and negatives are sampled at random from the remaining candidates.
Table~\ref{tab:under-translation-soft} shows some examples of UT subtle negatives for en-fr.


\textbf{OT subtle negatives.} 
We did not get meaningful insertion candidates using multilingual BERT LM for insertion into target to generate OT errors.
Hence, we randomly omitted up to five tokens incrementally from the source.
$40\%$ of the candidates are filtered out similar to the UT above using similarity scores derived from Glove embeddings and actual OT subtle negatives are sampled at random from the rest.
Table~\ref{tab:over-translation-soft} shows some examples of $(s, s', t)$ for en-fr. 



\textbf{Gross negatives.}
Subtle negatives are errors with a few independent tokens added or omitted. However, some errors are off by complete phrases or sentences. These errors are synthesized by removing or adding complete sentences to the source. 
Some gross examples are given in Table~\ref{tab:hard-candidates} for en-fr.

We generated a total of over 6.1 million subtitle pairs covering 28 target languages, namely, Arabic, Bengali, Chinese (Simplified), Chinese (Traditional), Czech, Danish, Dutch, Finnish, French, German, Greek, Hebrew, Hindi, Indonesian, Italian, Korean, Marathi, Norwegian, Polish, Portuguese, Romanian, Russian, Spanish, Swedish, Tamil, Telugu, Thai, Turkish. 
Subtle errors constitute 83\% of all errors and 17\% are gross errors. 30\% of all samples are OT errors, another 30\% are UT errors  and remaining 40\% have no errors. 
Dataset was split 80-20 into train-validation sets.

\textbf{Evaluation data.}
We created a high-quality human annotated dataset for evaluation. 
A random sample of 5,000 subtitle pairs each for English as source language and 8 target languages (Chinese (Traditional), French, Hindi, Italian, Portuguese, Russian, Spanish, Turkish). We carefully chose our target language set to include as many different scripts as possible. We asked three professional subtitle translators to mark if each pair had OT error or UT error along with the error-cause. They were asked to not mark anything if there was any confusion. We collated the results and selected sentences where there was an \textit{unanimous agreement} of either it having error or not having error. Out of 40,000 pairs, there was an unanimous agreement on 30,942 pairs; with 30,003 marked NE, 307 marked OT and 632 marked UT. Table~\ref{tab:human-validation-results} shows language-wise distribution of this dataset.

\section{Model Training and Results}

We trained models that classify each sample to one of the three classes; NE (No-Error), OT (Over-Translation) and UT (Under-Translation). 
A single model is trained for all language pairs with Adam optimizer~\cite{DBLP:journals/corr/KingmaB14} and a multilingual cased BERT model \cite{DBLP:conf/naacl/DevlinCLT19} to generate token embeddings.
We experimented with four model architectures (1) \textbf{Weighted GRU} which was a GRU network with re-scaled weights. (2) \textbf{(M1) GRU + CNN} which had three CNN layers on one GRU encoder. (3) \textbf{(M2) CNN} with four CNN layers. (4) \textbf{Hybrid M1 + M2} was a combined architecture of M1 and M2. We gave the input BERT embeddings to a GRU + CNN network and a vanilla CNN network; concatenated the embeddings of both parallel networks into common linear layer, followed by a softmax. 
Table~\ref{tab:model-comparison-results} tabulates their performances. \textbf{M1} was performing better in classifying UT class while \textbf{M2} was better classifying NE class; therefore we combined both the architectures to train the best performing model. 

\begin{table}[htbp]
  \centering
  \caption{Training and validation accuracy are reported on synthetic datasets. Test accuracy and weighted F1 scores are on human annotated dataset for all four model architectures.}
  \resizebox{\columnwidth}{!}{
    \begin{tabular}{c|c|c|c|c}
    \multirow{2}{*}{\textbf{Model Name}} & \textbf{Train} & \textbf{Validation} & \multicolumn{2}{c}{\textbf{Test}} \\ \cline{4-5}
     & \textbf{Accuracy} & \textbf{Accuracy} & \textbf{Accuracy} & \textbf{F1} \\ \hline
    \textbf{Weighted GRU} & 0.8227 & 0.8204 & 0.7703 & 0.8478 \\
    \textbf{(M1) GRU + CNN} & 0.7519 & 0.7512 & 0.6649 & 0.7762 \\
    \textbf{(M2) CNN} & \textbf{0.9233} & \textbf{0.9150} & 0.8609 & 0.9024 \\
    \textbf{Hybrid M1 + M2} & 0.9058 & 0.8983 & \textbf{0.8932} & \textbf{0.9194} \\
    \end{tabular}%
  }
  \label{tab:model-comparison-results}%
\end{table}%

\begin{table}[htbp]
  \centering
  \caption{Language-wise Human Annotated Test Dataset results for best performing model. NE: No-error class; OT: Over-translation class; UT: Under-translation}
  \resizebox{\columnwidth}{!}{
    \begin{tabular}{c|c|c|c|c}
    \textbf{Target} & \multirow{2}{*}{\textbf{(\#NE, \#UT, \#OT)}} & \multirow{2}{*}{\textbf{Accuracy}} & \multirow{2}{*}{\textbf{F1}} & \textbf{Error} \\ 
    \textbf{Language} & & & & \textbf{Recall} \\ \hline
     es       &  (4030, 46, 70)   & 0.8567 & 0.8986 & 0.3190 \\
     fr       &  (4181, 7, 11)    & 0.8183 & 0.8961 & 0.3333 \\
     hi       &  (3696, 25, 61)   & 0.9535 & 0.9562 & 0.1512 \\
     it       &  (3430, 62, 89)   & 0.8802 & 0.8998 & 0.1921 \\
     pt       &  (2811, 34, 145)  & 0.8793 & 0.8920 & 0.3575 \\
     ru       &  (4104, 35, 121)  & 0.9087 & 0.9213 & 0.2051 \\
     tr       &  (3573, 73, 93)   & 0.8791 & 0.8969 & 0.1627 \\
     zh-hant  &  (4178, 25, 42)   & 0.8985 & 0.9324 & 0.3134 \\

    \end{tabular}%
  }
  \label{tab:human-validation-results}%
\end{table}%

Table~\ref{tab:human-validation-results} shows language-wise results on evaluation data. 
The error recall captures the percentage of OT/UT errors that are flagged by the system for further manual processing.
Depending on the language, the model can capture only about 15-35\% of the errors by learning solely from synthesized errors.
This stresses on the need to improve methods for synthesizing errors closer to the actual distribution. 
However, this is very tricky since there is subjectivity involved in judging adequacy due to the complexities of paraphrasing as shown by the large disagreement (on about 25\% of the samples) among the three voters in human annotation.

\section{Conclusion}
We presented a novel approach of detecting over-translation (OT) and under-translation (UT) in human translated subtitle data using BERT.
We pursue this as an alternative direction to HTER predicting QE methods popular for MT output analysis.
As an advantage this extends to any translations, not necessarily MT, while explicitly modelling fluency and adequacy that helps translators.
Even though our training dataset mostly comprised of synthetically introduced errors, it performs well on high-quality human annotated data. We wish to improve upon this in two directions; (1) by improving error patterns through tighter coupling with human translators following methods similar to~\cite{DBLP:conf/iclr/KaushikHL20} and~\cite{DBLP:conf/emnlp/0001ABBBCDDEGGH20} and (2) by focusing on localizing errors down to tokens within the sentences rather than merely flagging the whole sentence. For OT, we could identify the sequence of tokens in target subtitle that cause the issue and similarly in source subtitle for UT. 

\bibliography{acl2020}

\begin{thebibliography}{23}
\expandafter\ifx\csname natexlab\endcsname\relax\def\natexlab#1{#1}\fi

\bibitem[{Artetxe and Schwenk(2019)}]{DBLP:journals/tacl/ArtetxeS19}
Mikel Artetxe and Holger Schwenk. 2019.
\newblock \href {https://transacl.org/ojs/index.php/tacl/article/view/1742}
  {Massively multilingual sentence embeddings for zero-shot cross-lingual
  transfer and beyond}.
\newblock \emph{Trans. Assoc. Comput. Linguistics}, 7:597--610.

\bibitem[{Blatz et~al.(2004)Blatz, Fitzgerald, Foster, Gandrabur, Goutte,
  Kulesza, Sanch{\'{\i}}s, and Ueffing}]{DBLP:conf/coling/BlatzFFGGKSU04}
John Blatz, Erin Fitzgerald, George~F. Foster, Simona Gandrabur, Cyril Goutte,
  Alex Kulesza, Alberto Sanch{\'{\i}}s, and Nicola Ueffing. 2004.
\newblock \href {https://www.aclweb.org/anthology/C04-1046/} {Confidence
  estimation for machine translation}.
\newblock In \emph{{COLING} 2004, 20th International Conference on
  Computational Linguistics, Proceedings of the Conference, 23-27 August 2004,
  Geneva, Switzerland}.

\bibitem[{Devlin et~al.(2019)Devlin, Chang, Lee, and
  Toutanova}]{DBLP:conf/naacl/DevlinCLT19}
Jacob Devlin, Ming{-}Wei Chang, Kenton Lee, and Kristina Toutanova. 2019.
\newblock \href {https://doi.org/10.18653/v1/n19-1423} {{BERT:} pre-training of
  deep bidirectional transformers for language understanding}.
\newblock In \emph{Proceedings of the 2019 Conference of the North American
  Chapter of the Association for Computational Linguistics: Human Language
  Technologies, {NAACL-HLT} 2019, Minneapolis, MN, USA, June 2-7, 2019, Volume
  1 (Long and Short Papers)}, pages 4171--4186. Association for Computational
  Linguistics.

\bibitem[{Fonseca et~al.(2019)Fonseca, Yankovskaya, Martins, Fishel, and
  Federmann}]{DBLP:conf/wmt/FonsecaYMFF19}
Erick~R. Fonseca, Lisa Yankovskaya, Andr{\'{e}} F.~T. Martins, Mark Fishel, and
  Christian Federmann. 2019.
\newblock \href {https://doi.org/10.18653/v1/w19-5401} {Findings of the {WMT}
  2019 shared tasks on quality estimation}.
\newblock In \emph{Proceedings of the Fourth Conference on Machine Translation,
  {WMT} 2019, Florence, Italy, August 1-2, 2019 - Volume 3: Shared Task Papers,
  Day 2}, pages 1--10. Association for Computational Linguistics.

\bibitem[{Gardner et~al.(2020)Gardner, Artzi, Basmova, Berant, Bogin, Chen,
  Dasigi, Dua, Elazar, Gottumukkala, Gupta, Hajishirzi, Ilharco, Khashabi, Lin,
  Liu, Liu, Mulcaire, Ning, Singh, Smith, Subramanian, Tsarfaty, Wallace,
  Zhang, and Zhou}]{DBLP:conf/emnlp/0001ABBBCDDEGGH20}
Matt Gardner, Yoav Artzi, Victoria Basmova, Jonathan Berant, Ben Bogin, Sihao
  Chen, Pradeep Dasigi, Dheeru Dua, Yanai Elazar, Ananth Gottumukkala, Nitish
  Gupta, Hannaneh Hajishirzi, Gabriel Ilharco, Daniel Khashabi, Kevin Lin,
  Jiangming Liu, Nelson~F. Liu, Phoebe Mulcaire, Qiang Ning, Sameer Singh,
  Noah~A. Smith, Sanjay Subramanian, Reut Tsarfaty, Eric Wallace, Ally Zhang,
  and Ben Zhou. 2020.
\newblock \href {https://www.aclweb.org/anthology/2020.findings-emnlp.117/}
  {Evaluating models' local decision boundaries via contrast sets}.
\newblock In \emph{Proceedings of the 2020 Conference on Empirical Methods in
  Natural Language Processing: Findings, {EMNLP} 2020, Online Event, 16-20
  November 2020}, pages 1307--1323. Association for Computational Linguistics.

\bibitem[{Gupta and Nelakanti(2020)}]{DBLP:journals/corr/abs-2004-13828}
Prabhakar Gupta and Anil Nelakanti. 2020.
\newblock \href {http://arxiv.org/abs/2004.13828} {Deepsubqe: Quality
  estimation for subtitle translations}.
\newblock \emph{CoRR}, abs/2004.13828.

\bibitem[{Gupta et~al.(2019)Gupta, Sharma, Pitale, and
  Kumar}]{DBLP:journals/corr/abs-1909-05362}
Prabhakar Gupta, Mayank Sharma, Kartik Pitale, and Keshav Kumar. 2019.
\newblock \href {http://arxiv.org/abs/1909.05362} {Problems with automating
  translation of movie/tv show subtitles}.
\newblock \emph{CoRR}, abs/1909.05362.

\bibitem[{Kaneko et~al.(2020)Kaneko, Mita, Kiyono, Suzuki, and
  Inui}]{DBLP:conf/acl/KanekoMKSI20}
Masahiro Kaneko, Masato Mita, Shun Kiyono, Jun Suzuki, and Kentaro Inui. 2020.
\newblock \href {https://www.aclweb.org/anthology/2020.acl-main.391/}
  {Encoder-decoder models can benefit from pre-trained masked language models
  in grammatical error correction}.
\newblock In \emph{Proceedings of the 58th Annual Meeting of the Association
  for Computational Linguistics, {ACL} 2020, Online, July 5-10, 2020}, pages
  4248--4254. Association for Computational Linguistics.

\bibitem[{Kaushik et~al.(2020)Kaushik, Hovy, and
  Lipton}]{DBLP:conf/iclr/KaushikHL20}
Divyansh Kaushik, Eduard~H. Hovy, and Zachary~Chase Lipton. 2020.
\newblock \href {https://openreview.net/forum?id=Sklgs0NFvr} {Learning the
  difference that makes {A} difference with counterfactually-augmented data}.
\newblock In \emph{8th International Conference on Learning Representations,
  {ICLR} 2020, Addis Ababa, Ethiopia, April 26-30, 2020}. OpenReview.net.

\bibitem[{Kepler et~al.(2019)Kepler, Trénous, Treviso, Vera, and
  Martins}]{openkiwi}
Fábio Kepler, Jonay Trénous, Marcos Treviso, Miguel Vera, and André F.~T.
  Martins. 2019.
\newblock \href {https://www.aclweb.org/anthology/P19-3020} {Open{K}iwi: An
  open source framework for quality estimation}.
\newblock In \emph{Proceedings of the 57th Annual Meeting of the Association
  for Computational Linguistics--System Demonstrations}, pages 117--122,
  Florence, Italy. Association for Computational Linguistics.

\bibitem[{Kingma and Ba(2015)}]{DBLP:journals/corr/KingmaB14}
Diederik~P. Kingma and Jimmy Ba. 2015.
\newblock \href {http://arxiv.org/abs/1412.6980} {Adam: {A} method for
  stochastic optimization}.
\newblock In \emph{3rd International Conference on Learning Representations,
  {ICLR} 2015, San Diego, CA, USA, May 7-9, 2015, Conference Track
  Proceedings}.

\bibitem[{Lichtarge et~al.(2020)Lichtarge, Alberti, and
  Kumar}]{DBLP:journals/tacl/LichtargeAK20}
Jared Lichtarge, Chris Alberti, and Shankar Kumar. 2020.
\newblock \href {https://transacl.org/ojs/index.php/tacl/article/view/2047}
  {Data weighted training strategies for grammatical error correction}.
\newblock \emph{Trans. Assoc. Comput. Linguistics}, 8:634--646.

\bibitem[{Omelianchuk et~al.(2020)Omelianchuk, Atrasevych, Chernodub, and
  Skurzhanskyi}]{DBLP:conf/bea/OmelianchukACS20}
Kostiantyn Omelianchuk, Vitaliy Atrasevych, Artem~N. Chernodub, and Oleksandr
  Skurzhanskyi. 2020.
\newblock \href {https://www.aclweb.org/anthology/2020.bea-1.16/} {Gector -
  grammatical error correction: Tag, not rewrite}.
\newblock In \emph{Proceedings of the Fifteenth Workshop on Innovative Use of
  {NLP} for Building Educational Applications, BEA@ACL 2020, Online, July 10,
  2020}, pages 163--170. Association for Computational Linguistics.

\bibitem[{Papineni et~al.(2002)Papineni, Roukos, Ward, and
  Zhu}]{DBLP:conf/acl/PapineniRWZ02}
Kishore Papineni, Salim Roukos, Todd Ward, and Wei{-}Jing Zhu. 2002.
\newblock \href {https://doi.org/10.3115/1073083.1073135} {Bleu: a method for
  automatic evaluation of machine translation}.
\newblock In \emph{Proceedings of the 40th Annual Meeting of the Association
  for Computational Linguistics, July 6-12, 2002, Philadelphia, PA, {USA}},
  pages 311--318. {ACL}.

\bibitem[{Pennington et~al.(2014)Pennington, Socher, and
  Manning}]{DBLP:conf/emnlp/PenningtonSM14}
Jeffrey Pennington, Richard Socher, and Christopher~D. Manning. 2014.
\newblock \href {https://doi.org/10.3115/v1/d14-1162} {Glove: Global vectors
  for word representation}.
\newblock In \emph{Proceedings of the 2014 Conference on Empirical Methods in
  Natural Language Processing, {EMNLP} 2014, October 25-29, 2014, Doha, Qatar,
  {A} meeting of SIGDAT, a Special Interest Group of the {ACL}}, pages
  1532--1543. {ACL}.

\bibitem[{Poirier(2014)}]{DBLP:journals/jides/Poirier14}
{\'{E}}ric~Andr{\'{e}} Poirier. 2014.
\newblock \href {https://doi.org/10.1016/j.jides.2015.02.004} {A method for
  automatic detection and manual localization of content-based translation
  errors and shifts}.
\newblock \emph{J. Innov. Digit. Ecosyst.}, 1(1-2):38--46.

\bibitem[{Popovi{\'c} and Ney(2011)}]{popovic2011towards}
Maja Popovi{\'c} and Hermann Ney. 2011.
\newblock Towards automatic error analysis of machine translation output.
\newblock \emph{Computational Linguistics}, 37(4):657--688.

\bibitem[{Snover et~al.(2006)Snover, Dorr, Schwartz, and
  Micciulla}]{Snover2006ASO}
Matthew Snover, Bonnie~J. Dorr, Richard Schwartz, and Linnea Micciulla. 2006.
\newblock A study of translation edit rate with targeted human annotation.

\bibitem[{Specia et~al.(2009)Specia, Turchi, Cancedda, Cristianini, and
  Dymetman}]{DBLP:conf/eamt/SpeciaTCCD09}
Lucia Specia, Marco Turchi, Nicola Cancedda, Nello Cristianini, and Marc
  Dymetman. 2009.
\newblock \href {https://www.aclweb.org/anthology/2009.eamt-1.5/} {Estimating
  the sentence-level quality of machine translation systems}.
\newblock In \emph{Proceedings of the 13th Annual conference of the European
  Association for Machine Translation, {EAMT} 2009, Barcelona, Spain, Map
  14-15, 2009}. European Association for Machine Translation.

\bibitem[{Sun et~al.(2020)Sun, Guzm{\'{a}}n, and
  Specia}]{DBLP:conf/acl/SunGS20}
Shuo Sun, Francisco Guzm{\'{a}}n, and Lucia Specia. 2020.
\newblock \href {https://www.aclweb.org/anthology/2020.acl-main.558/} {Are we
  estimating or guesstimating translation quality?}
\newblock In \emph{Proceedings of the 58th Annual Meeting of the Association
  for Computational Linguistics, {ACL} 2020, Online, July 5-10, 2020}, pages
  6262--6267. Association for Computational Linguistics.

\bibitem[{Tu et~al.(2016)Tu, Lu, Liu, Liu, and Li}]{DBLP:conf/acl/TuLLLL16}
Zhaopeng Tu, Zhengdong Lu, Yang Liu, Xiaohua Liu, and Hang Li. 2016.
\newblock \href {https://doi.org/10.18653/v1/p16-1008} {Modeling coverage for
  neural machine translation}.
\newblock In \emph{Proceedings of the 54th Annual Meeting of the Association
  for Computational Linguistics, {ACL} 2016, August 7-12, 2016, Berlin,
  Germany, Volume 1: Long Papers}. The Association for Computer Linguistics.

\bibitem[{Wu and Dredze(2019)}]{DBLP:conf/emnlp/WuD19}
Shijie Wu and Mark Dredze. 2019.
\newblock \href {https://doi.org/10.18653/v1/D19-1077} {Beto, bentz, becas: The
  surprising cross-lingual effectiveness of {BERT}}.
\newblock In \emph{Proceedings of the 2019 Conference on Empirical Methods in
  Natural Language Processing and the 9th International Joint Conference on
  Natural Language Processing, {EMNLP-IJCNLP} 2019, Hong Kong, China, November
  3-7, 2019}, pages 833--844. Association for Computational Linguistics.

\bibitem[{Yang et~al.(2018)Yang, Zhang, Qin, Zhang, Lin, and
  Su}]{DBLP:conf/nlpcc/YangZQZLS18}
Jing Yang, Biao Zhang, Yue Qin, Xiangwen Zhang, Qian Lin, and Jinsong Su. 2018.
\newblock \href {https://doi.org/10.1007/978-3-319-99495-6\_25} {Otem{\&}utem:
  Over- and under-translation evaluation metric for {NMT}}.
\newblock In \emph{Natural Language Processing and Chinese Computing - 7th
  {CCF} International Conference, {NLPCC} 2018, Hohhot, China, August 26-30,
  2018, Proceedings, Part {I}}, volume 11108 of \emph{Lecture Notes in Computer
  Science}, pages 291--302. Springer.

\end{thebibliography}
\bibliographystyle{acl_natbib}

\end{document}